\documentclass[10pt,twocolumn,letterpaper]{article}

\usepackage{lineno}
\usepackage{cvpr}              % To produce the CAMERA-READY version
\usepackage[accsupp]{axessibility}
\usepackage{graphicx}
\usepackage{wrapfig}
\usepackage{amsfonts}
\usepackage{amsmath}
\usepackage{color}
\usepackage{booktabs}
\usepackage{multirow}
\usepackage[normalem]{ulem}
\useunder{\uline}{\ul}{}
\usepackage{array}
\usepackage{colortbl}
\usepackage{xcolor}
\usepackage{caption}
\newtheorem{theorem}{Theorem}
\usepackage{bbding}

\definecolor{cvprblue}{rgb}{0.21,0.49,0.74}
\usepackage[pagebackref,breaklinks,colorlinks,allcolors=cvprblue]{hyperref}

\def\paperID{5942} % *** Enter the Paper ID here
\def\confName{CVPR}
\def\confYear{2025}

\title{Towards Improved Text-Aligned Codebook Learning: \\
Multi-Hierarchical Codebook-Text Alignment with Long Text}

\author{Guotao Liang\textsuperscript{1,2}, Baoquan Zhang\textsuperscript{1}\thanks{Corresponding Authors}, Zhiyuan Wen\textsuperscript{2}, Junteng Zhao\textsuperscript{1}, Yunming Ye\textsuperscript{1}, Kola Ye\textsuperscript{3}, Yao He\textsuperscript{3} \\
  \textsuperscript{1}Harbin Institute of Technology, Shenzhen \textsuperscript{2}Peng Cheng Laboratory,
    \textsuperscript{3}SiFar Company \\ 
    \tt\small lianggt@pcl.ac.cn, \{baoquanzhang, yeyunming\}@hit.edu.cn, 23S051042@stu.hit.edu.cn, \\
    \tt\small \{wenzhiyuan2012, kolaygm, heyao18818\}@gmail.com
}

\begin{document}
\maketitle
\begin{abstract}
% what is VQ
Image quantization is a crucial technique in image generation, aimed at learning a codebook that encodes an image into a discrete token sequence. 
Recent advancements have seen researchers exploring learning multi-modal codebook (i.e., text-aligned codebook) by utilizing image caption semantics, aiming to enhance codebook performance in cross-modal tasks. 
However, existing image-text paired datasets exhibit a notable flaw in that the text descriptions tend to be overly concise, failing to adequately describe the images and provide sufficient semantic knowledge, resulting in limited alignment of text and codebook at a fine-grained level.
In this paper, we propose a novel Text-Augmented Codebook Learning framework, named TA-VQ, which generates longer text for each image using the visual-language model for improved text-aligned codebook learning.
However, the long text presents two key challenges: how to encode text and how to align codebook and text. To tackle two challenges, we propose to split the long text into multiple granularities for encoding, i.e., word, phrase, and sentence, so that the long text can be fully encoded without losing any key semantic knowledge. Following this, a hierarchical encoder and novel sampling-based alignment strategy are designed to achieve fine-grained codebook-text alignment. Additionally, our method can be seamlessly integrated into existing VQ models. Extensive experiments in reconstruction and various downstream tasks demonstrate its effectiveness compared to previous state-of-the-art approaches. 

\end{abstract}    
\section{Introduction}
\label{sec:intro}
In recent years, vector quantization (VQ)-based image modeling has emerged as a prominent technique in the field of image generation \cite{van2017vqvae}. VQ-based image modeling aims to encode continuous images into discrete code sequences using a trainable visual codebook within an encoder-quantizer-decoder framework. Under this paradigm, the learned codebook, capturing rich image features, can be leveraged to generate high-fidelity images through auto-regression \cite{rombach2022high,Omar2019Image,esser2021taming} or discrete diffusion methods\cite{gu2022vqdiffusion}.

\begin{figure}[t]
  \centering
   \includegraphics[width=1.0\linewidth]{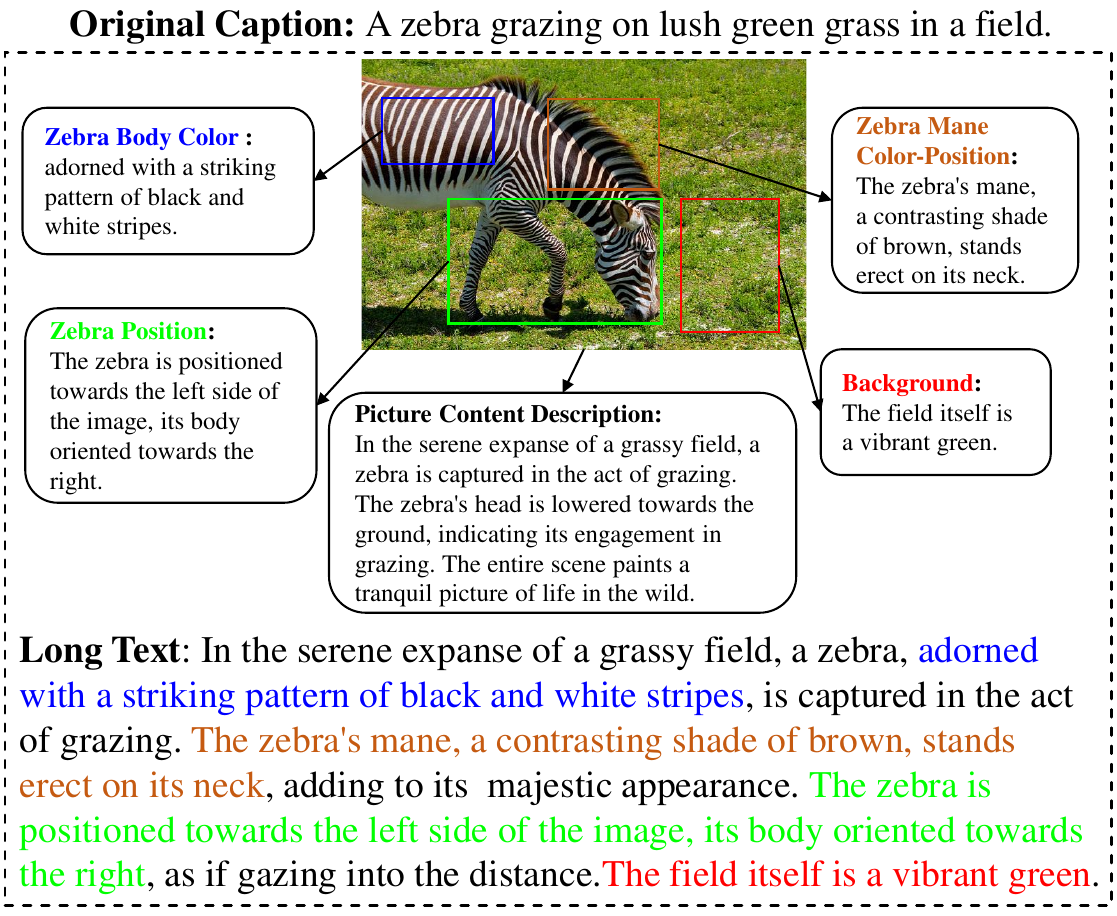}
   \caption{Example of comparing origin caption and long text. The original caption is brief and primarily provides a general description of the image. It lacks detail about the background and certain key elements within the image. In contrast, the long text can offer a more comprehensive image description, including zebra body color, position, mane color position, and background description. This additional text offers richer context, contributing to more robust and effective text-aligned codebook learning.}
   \label{fig:motivation}
   \vspace{-4mm}
\end{figure}

To further enhance the performance of VQ-based methods, several innovative techniques and ideas \cite{Lin23Catch, gu2022rethinking,liu2023learning} have been introduce to learn a more robust codebook representation, such as introducing adversarial loss \cite{esser2021taming}, learning compact code sequence \cite{huang2023towards,huang2023not}, or dealing with codebook collapse \cite{Zheng2023CVQ, Zhang2023Regularized,huh2023overcoming,Takida22Variational,zhang2024vqct}. Different from the aforementioned methods that only learn single-modal codebook, Liang \etal. \cite{liang2024lgvq} propose LG-VQ, which incorporates pre-trained text semantics to learn text-aligned codebook, thereby improving the performance of VQ-based models in various cross-modal downstream tasks. 

Despite the success of LG-VQ, it has limitations regarding insufficient alignment between codebook and text, primarily due to the brevity of the existing image captions. As illustrated in~\cref{fig:motivation}, we can see that the original caption is concise, focusing solely on the main object while omitting details about the background and other key elements of the image. This brevity results in a lack of sufficient semantic information, hindering the learning of a well text-aligned codebook. Inspired by recent visual-language models (VLMs) \cite{girdhar2023imagebind,zhu2023minigpt,chen2023sharegpt4v}, we propose utilizing VLMs to generate more detailed description for each image, as depicted in~\cref{fig:motivation}. Compared to the original caption, the augmented text provides a more comprehensive description of the image. Resorting to such longer text, richer semantic knowledge can be obtained to learn more robust codebook representation and achieve improved text-aligned codebook. 

However, the augmented text usually contains hundreds of words, making it challenging for pre-trained language models \cite{Alec2021clip, devlin2018bert} to capture and retain all key semantic information during encoding. Additionally, encoding and using such lengthy text imposes a significant computational burden. To address this issue, we propose splitting the long text into multiple granularities for encoding, \ie, word, phrase, and sentence. Its advantage is that the long text semantics can be fully encoded without losing key semantic knowledge by resorting to different granularities. Unfortunately, we are in another dilemma. The multi-granularity text semantics and single code sequence semantics are structurally inconsistent, making direct alignment more difficult. To handle this, we propose a hierarchical encoder that encodes images into multi-hierarchical code representations, each hierarchy corresponding to a specific semantic granularity text. This design ensures a more precise and structurally consistent alignment relationship between image codes and multi-granularity text semantics. Based on the consistent alignment relationship, we propose a novel sampling-based alignment strategy, which can achieve fine-grained codebook-text alignment while not introducing much computational overhead.
As a result, a novel text-augmented codebook learning framework, termed \textit{Text-Augmented VQ} (TA-VQ), is proposed for improved text-aligned codebook learning.

In a nutshell, we summarize our main contributions as:
\begin{itemize}
    \item We propose a novel text-augmented codebook learning framework, TA-VQ, which leverages VLMs to generate longer text for each image, improving text-aligned codebook learning. This approach enables 1) fine-grained alignment through detailed image descriptions, and 2) richer semantic knowledge for robust codebook learning. 
    \item We propose to encode long text semantics at three granularities, \ie, word, phrase, and sentence, aiming to retain as much key semantic knowledge as possible. Resorting to this multi-granularity text semantics, a hierarchical encoder and novel sampling-based alignment strategy are designed to achieve fine-grained codebook-text alignment.
    \item We conduct comprehensive experiments on three public datasets, demonstrating that our method surpasses previous state-of-the-art models in reconstruction quality and performance across various downstream tasks.
\end{itemize}
\section{Related Works}
\subsection{Vector Quantization for Image Generation}
Vector quantization (VQ) plays a crucial role in image generation by compressing continuous images into discrete token sequences using a learnable codebook \cite{yu2021vitvq,chang2022maskgit,lee2022autoregressive,gu2022vqdiffusion}. Oord \etal.~\cite{van2017vqvae} introduced VQ-VAE, an encoder-quantizer-decoder architecture for vector quantization. However, VQ-VAE is limited in generating high-fidelity images due to the insufficient codebook representation. Subsequently, various methods \cite{esser2021taming,huang2023towards,huang2023not,Lin23Catch,liang2024lgvq,gu2024rethinking} have been proposed to enhance image generation quality. For instance, Esser \etal. \cite{esser2021taming} propose VQ-GAN, integrating adversarial and perceptual losses based on VQ-VAE, which significantly improves the quality of image generation. CVQ \cite{Zheng2023CVQ} addresses codebook collapse by selecting encoded features as anchors to update ``dead'' code representations. VQCT \cite{zhang2024vqct} introduces pre-trained word semantics as the initial codebook and then utilizes the semantic relationship of words to achieve cooperative optimization between codes to eliminate the codebook collapse issue. Recently, some researchers \cite{liu2023lqae,yu2023spae,zhu2024beyond} explore mapping the images to the word tokens of LLMs by viewing images as ``foreign languages'' to improve the performance of the codebook on downstream tasks with the help large language models (LLMs). Unlike the aforementioned methods that focus solely on single-modal codebook learning, Liang \etal. \cite{liang2024lgvq} propose a novel multi-modal codebook learning framework, called LG-VQ, by introducing pre-trained text semantics as supervised information, which significantly improves the performance of VQ-based models in various cross-modal downstream tasks. 

In this paper, we also focus on learning text-aligned codebook. Different from the existing methods that rely on limited image captions, our approach leverages the VLMs to extend the text description for each image and proposes a novel alignment method to efficiently encode and exploit the semantics of extended texts for improving text-aligned codebook learning.

\subsection{Visual-Language Models}
Inspired by the success of Large Language Models (LLMs) in the field of natural language processing \cite{brown2020gpt3,touvron2023llama,dubey2024llama3,raffel2020t5}, lots of researchers \cite{zhu2023minigpt,girdhar2023imagebind,chen2023sharegpt4v} have begun to explore integrating visual information into LLMs to form Visual Language Models (VLMs) for jointly solving various multi-modal tasks, such as visual grounding \cite{deng2021transvg}, visual question answering \cite{lu2016vqa}, image description \cite{mokady2021clipcap}, etc. There are currently two fusion strategies, \ie, token-level and feature-level fusion. The token-level refers to the visual features being transformed into tokens, which LLMs can understand, through a learnable Adapter Network, and then are concatenated with text tokens and fed into LLMs. This learnable Adapter Network includes Q-Former \cite{li2023blip2,Dai2023InstructBLIP,zhang2023videoLlama,chen2023xLLM} and MLP \cite{liu2024visual,pi2023detgpt,su2023pandagpt,zhang2023pmc}. The feature-level refers to inserting extra modules to enable text and visual features to interact more deeply, such as cross-attention layers \cite{alayrac2022flamingo}, visual expert module \cite{wang2023cogvlm}. These VLMs have shown powerful capabilities in handling various visual understanding tasks.

In this paper, we propose to employ VLMs to generate more detailed descriptions for each image. The advantage is that longer text can bring richer semantic knowledge for more robust codebook learning.

\section{Preliminaries}
\subsection{VQ-VAE}
\begin{figure*}
    \centering
    \includegraphics[width=1\linewidth]{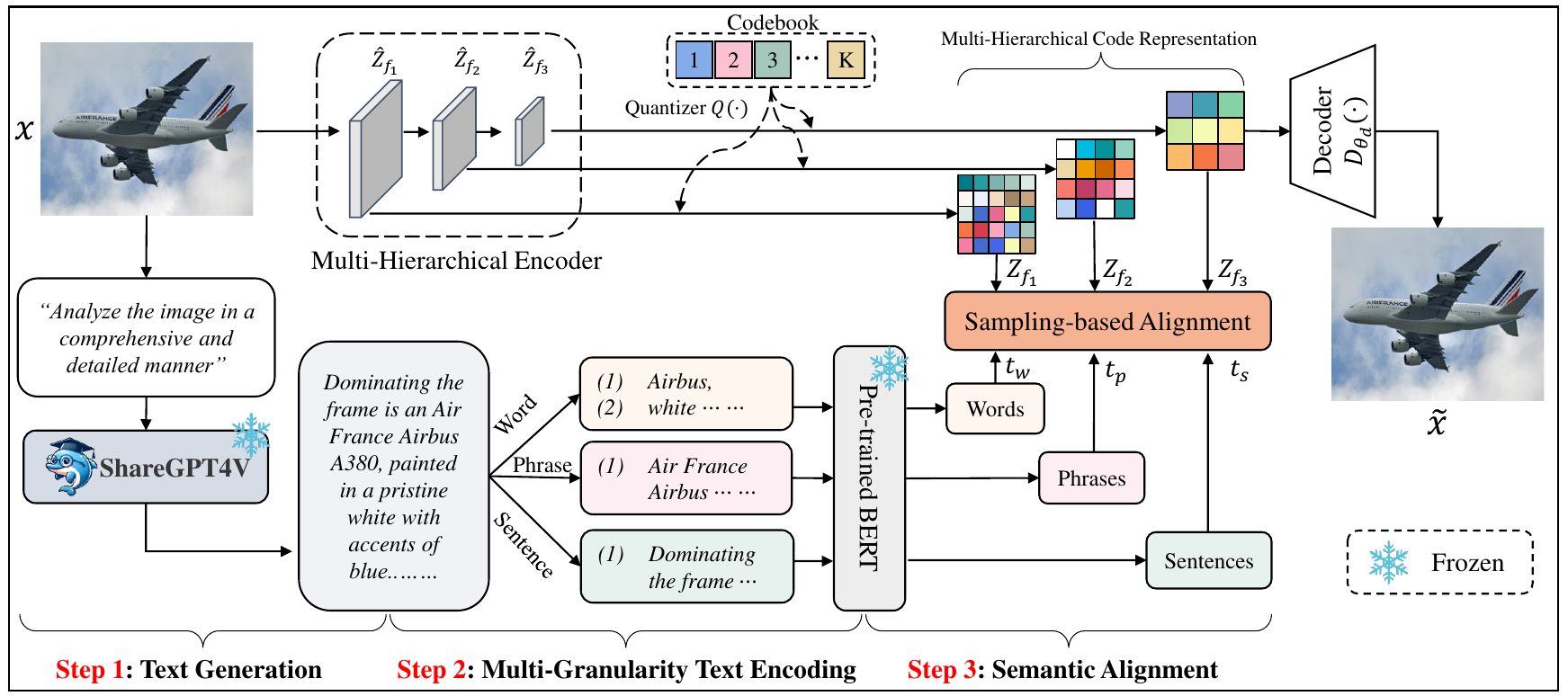}
    \caption{The illustration of our proposed TA-VQ framework. The image is first fed to VLM to generate a more detailed text description, and then the text is split into multiple granularities for encoding, \ie, word ($t_w$), phrase ($t_p$), and sentence semantics ($t_s$). Subsequently, the multi-hierarchical encoder encodes and quantizes the image into multi-hierarchical code representation, \ie, $Z_{f_1}$, $Z_{f_2}$, and $Z_{f_3}$. The sampling-base alignment module is employed to achieve $Z_{f_1}$, $Z_{f_2}$, $Z_{f_3}$ and $t_w$, $t_p$, $t_s$ alignment. Finally, the decoder is used to reconstruct the origin image using $Z_{f_3}$.}
    \label{fig:model}
    \vspace{-3mm}
\end{figure*}

VQ-VAE \cite{van2017vqvae} trains a visual codebook to encode the image into a discrete code sequence through an encoder-quantizer-decoder framework. As illustrated
in~\cref{fig:model}, the codebook is defined as $C=\{(k, e_k)\}_{k}^{K}$ which consists of learnable $K$ entries $e_k \in \mathbb{R}^{d_z}$ with dimension $d_z$. Given an input image $x \in \mathbb{R}^{H \times W \times C}$, where $H$, $W$, and $C$ represent the height, width, and channel of the image respectively. The encoder $E_{\theta_{e}}(\cdot)$ with parameter $\theta_{e}$ and down-sampling factor $f$, which gradually compress the original image into a grid feature $\hat{Z} = E_{\theta_{e}}(x) \in \mathbb{R}^{\frac{H}{f} \times \frac{W}{f} \times d_z}$. The quantizer $Q(\cdot)$ replaces each grid feature with a code representation based on the distance between $\hat{Z}$ and $C$. That is:
\begin{equation}
    z_i = Q(\hat{z}_{i}) = \mathop{\text{argmin}}\limits_{e_k \in \mathcal{Z}} \Vert\hat{z}_{i} - e_k\Vert.
\label{eq:VQquantizer}
\end{equation} 
As a result, quantized code representation $Z$ can be obtained, the decoder $D_{\theta_{d}}(\cdot)$ with parameter $\theta_{d}$ is employed to reconstruct the input image by $\widetilde{x} = D_{\theta_{d}}(Z)$. The whole network is optimized by minimizing the following objective:
\begin{equation}
\mathcal{L}_{vq} = \Vert x - \widetilde{x}\Vert_2^2 + \Vert sg[E_{\theta_{e}}(x)] - Z \Vert_2^2 + \beta \Vert E_{\theta_{e}}(x) - sg[Z] \Vert_2^2,
\label{eq:vqloss}
\end{equation}
where $\beta$ is a hyper-parameter, $sg[\cdot]$ denotes the stop-gradient operator. The first term is reconstruction loss, the second is codebook loss, which encourages the codebook to be close grid features, and the third is the ``commitment loss'' \cite{van2017vqvae}. Recently, some researchers introduce pre-trained text semantics to learn text-aligned codebook. However, such concise text fails to provide sufficient semantic knowledge, resulting in insufficient codebook-text alignment.  

\subsection{Wasserstein distance} \label{sec:wd}
The Wasserstein distance \cite{frogner2015wass,yang2024improving}, known as Earth Mover’s Distance (EMD), measures the distance between two probability distributions or sets of weighted objects. The computation of the Wasserstein distance relies on solving the well-known transportation problem \cite{hitchcock1941distribution}. Specifically, suppose we have a set of suppliers $H = \{h_1, h_2, \dots, h_n\}$, where $h_i$ denotes supply good capacity of the $i$-th supplier. These suppliers need to provide goods to multiple consumers, and the quantity of goods required by each consumer is fixed $Q = \{q_1, q_2, \dots, q_m\}$, where $q_j$ indicates the amount needed by the $j$-th consumer. The cost of transporting one good unit from supplier $i$ to consumer $j$ is represented by $\gamma_{ij}$. The transportation problem is to find a least-expensive flow of goods $F = \{f_{ij} | i=1, \dots, n,j=1, \dots, m\}$ from the suppliers to the consumers that satisfy the consumers' demand:
\begin{equation}
\begin{aligned}
    \mathop{\text{minimize}}\limits_{f_{ij}} \quad & \sum_{i=1}^{n} \sum_{j=1}^{m} \gamma_{ij} f_{ij} \\
    \text{subject to} \quad & f_{ij} \geq 0, i = 1, \dots, n,j = 1, \dots, m \\
                            & \sum\nolimits_{j=1}^m f_{ij} = h_i, \quad i = 1, \dots, n \\
                            & \sum\nolimits_{i=1}^n f_{ij} = q_j, \quad j = 1, \dots, m.                    
\end{aligned}  
\label{eq:EMD}
\end{equation}
We can solve the Linear Programming problem or use some approximate algorithms \cite{jonker1988shortest,fan2017point,cuturi2013lightspeed} to obtain the optimal flow of goods $F$.
In our paper, we adopt Wasserstein distance to achieve the alignment between codebook and text.

\section{Proposed Method: TA-VQ}
Existing works attempt to improve codebook performance in downstream tasks by learning text-aligned codebook. However, these methods overlook the limitation of overly concise text descriptions (see ~\cref{fig:motivation} for more details), which provide insufficient semantic information for codebook-text alignment.
To address this, we propose TA-VQ, a novel codebook learning framework that generates longer text to improve codebook-text alignment. As illustrated in~\cref{fig:model}, our TA-VQ consists of three key steps, \ie, text generation, multi-granularity text encoding, and semantic alignment. Here, we elaborate on each step in order:

\textbf{Step 1: \emph{Text Generation.}} Different from existing works that adopt limiting text semantics, we employ ShareGPT4V~\cite{chen2023sharegpt4v} to generate longer and more detailed text for each image, improving text-codebook alignment. Specifically, following the ShareGPT4V, we construct a generic prompt that describes the image: \textit{``Analyze the image in a comprehensive and detailed manner''}. Then we feed the image and constructed prompt into ShareGPT4V to generate extended text descriptions. Notably, the same prompt is uniformly applied to all images.

\textbf{Step 2: \emph{Multi-Granularity Text Encoding.}} Although the generated long text can provide richer semantic knowledge, it also brings a key challenge, that is, how to effectively encode the text. To address this, we propose to split the long text into multiple granularities for encoding, \ie word, phrase, and sentence. Specifically, given a long text $t$, we employ a text process tool (\ie, TextBlob \cite{loria2018textblob}) to extract all sentences and phrases. Following this, the BERT language model \cite{devlin2018bert} is employed to encode them. As a result, the sentence and phrase semantics can be obtained, which is denoted by $t_{s} = \{s_1, s_2,\cdots,s_{\vert t_s \vert}\}$, where $s_i \in \mathbb{R}^{d} $ denotes $i$-th sentence representation with dimension $d$, and $t_{p} = \{p_1, p_2,\cdots,p_{\vert t_p \vert}\}$, where $p_i \in \mathbb{R}^{d}$ denotes $i$-th phrase representation, respectively. For word granularity, inspired by \cite{zhang2024vqct}, we focus on extracting visual-relevant words, such as nouns, adjectives, and quantifiers, rather than encoding all words. As a result, the word semantic can be obtained by BERT pre-trained word embedding, which is denoted by $t_{w} = \{w_1, w_2,\cdots,w_{\vert t_w \vert}\}$, where $w_i \in \mathbb{R}^{d}$ denotes $i$-th word representation. Resorting to text semantics of different granularities, the long text can be effectively encoded without losing key semantic knowledge, which is beneficial for achieving fine-grained alignment.

\textbf{Step 3: \emph{Semantic Alignment.}} As previously discussed, structurally inconsistent alignment relationships exist between single code sequence semantics and multi-granularity text semantics, making it difficult to align directly. To solve this, as shown in~\cref{fig:model}, we propose a hierarchical encoder that encodes and quantizes images into multi-hierarchical code representation, where each hierarchy corresponds to a specific granularity text semantic. This design leverages the inherent hierarchical structure of image encoding, where lower layers model basic semantics and deeper layers encode more abstract semantic features. This hierarchical structure mirrors the semantic hierarchical structure from word to phrase to sentence in the natural language. The key advantage of this design is that it enables a more precise and structurally consistent alignment relationship between image codes and multi-granularity text semantics. Formally, given an image, we first encode it into three hierarchical grid features, denoted as $\hat{Z}_{h} = \{\hat{Z}_{f_1}, \hat{Z}_{f_2}, \hat{Z}_{f_3}\}$, where $\hat{Z}_{f_j} \in \mathbb{R}^{\frac{H}{f_j} \times \frac{W}{f_j} \times d_z}$ and $f_j \in \{4,8,16\}$. Here, the $\hat{Z}_{f_1}$ captures the feature from the lower layers, while $\hat{Z}_{f_3}$ represents the feature from deeper layers. As a result, the multi-hierarchical image code representation can be obtained by~\cref{eq:VQquantizer}, which is denoted by $Z_{h} = \{Z_{f_1}, Z_{f_2}, Z_{f_3}\}$. Each hierarchical image code is then aligned with the corresponding granularity of text semantics (\ie, words, phrases, and sentences) based on a sampling-based alignment strategy. The details of the sampling-based alignment strategy will be elaborated in~\cref{sec:sampling}.

\subsection{Sampling-based Alignment Strategy} \label{sec:sampling}
In this subsection, we introduce how to align multi-hierarchical image code representation (\ie, $Z_{f_1}$, $Z_{f_2}$, $Z_{f_3}$) and multi-granularity text semantics (\ie, $t_w$, $t_p$, $t_s$). Since the alignment process is consistent across different granularities, we use the alignment between the $Z_{f_3}$ and sentences $t_s$ as an illustration. 

Actually, $Z_{f_3}$ and $t_s$ have unequal numbers (\ie, $\vert Z_{f_3} \vert \neq  \vert t_s \vert$), and there is no explicit alignment relationship in the dataset, making it impossible to measure the distance between them for optimization directly. One straightforward approach might involve computing the distance based on the mean representation of them, but this method overlooks critical local semantic information. To solve this problem, we formulate the semantic alignment problem as an optimal transport problem \cite{villani2008optimal}, \ie, minimizing the transport cost from codebook representation to text representation to achieve codebook-text semantic alignment. Following this idea, the Wasserstein distance is employed to calculate the optimal transport cost between $Z_{f_3}$ and $t_s$. According to the description in~\cref{sec:wd}, we can regard $Z_{f_3}$ as the suppliers, $t_s$ as the consumers, and the cost of transporting one good unit is measured by the pairwise distance between the $i$-th code representation and $j$-th sentence representation $\gamma_{ij} = \|z_i - s_j\|_2$. Following \cite{zhang2020deepemd}, we can embed the solution process of optimal flow $F$ into the neural network so that the transport cost can be minimized to achieve semantic alignment of codebook and text.

\begin{figure}[t]
    \centering
    \includegraphics[width=0.9\linewidth]{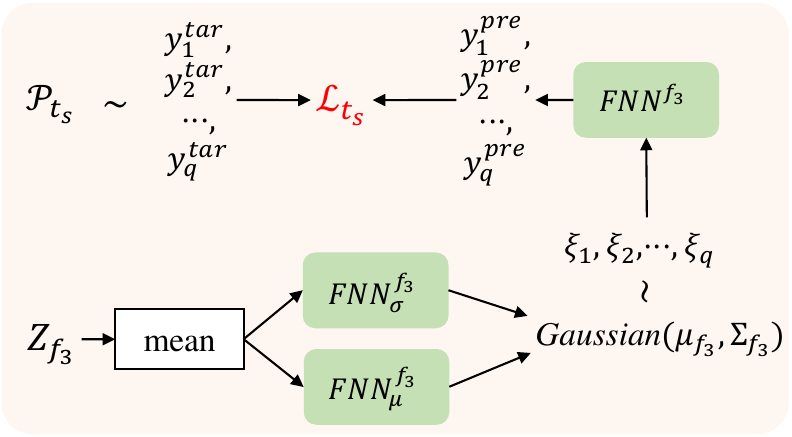}
    \caption{Illustration of the Sampling-based Alignment Strategy.}
    \label{fig:samplingalign}
    \vspace{-4mm}
\end{figure}

\textbf{Complexity Problem}. As mentioned above, considering the worst case, solving such an optimal flow $F$ has a complexity of $O(\max (\vert Z_{f_3} \vert^3, \vert t_s \vert^3))$. It is important to note that the computational burden arises primarily from the term $\vert Z_{f_3} \vert$. To address this challenge, we draw inspiration from the work of Lu \& Lu \cite{lu2020FFN_theorem}, which demonstrates that there exists a specialized feedforward neural network (FNN) so that the Wasserstein distance between a continuous distribution $\mu$ and a discrete distribution $\nu$ is small enough, as formally stated in~\cref{tr:FNN}. We regard $t_s$ as a discrete distribution and propose modeling $Z_{f_3}$ as a Gaussian Distribution, from which we can sample $\vert t_s \vert$ feature vectors using the reparameterization trick \cite{kingma2013reptick}. Then, we can use the sampled feature vectors to calculate the Wasserstein distance. Its advantage is that the computational complexity is reduced from $O(\vert Z_{f_3} \vert^3)$ to $O(\vert t_s \vert^3)$. To further reduce the computational burden, we propose to sample $q$ samples from $ t_s $ instead of using the entire $ t_s $. So that the computational burden is further reduced to $O(q^3)$. 
Specifically, as shown in~\cref{fig:samplingalign}, we equip FNNs to model $Z_{f_3}$ as a Gaussian Distribution:
\begin{equation}
    \begin{aligned}
        m_{f_3} &= \text{Mean} (Z_{f_3}), \\
        \mu_{f_3} &= \mathrm{FNN}_{u}^{f_3}(m_{f_3}), \\
        \Sigma_{f_3} &= \mathrm{diag}(\exp (\mathrm{FNN}_{\sigma}^{f_3} (m_{f_3})) ).
    \end{aligned}
\label{eq:guss}
\end{equation}
Then, according to the~\cref{tr:FNN}, we define the sentence information $t_s$ as a discrete distribution, denoted by $\mathcal{P}_{t_s} \triangleq \frac{1}{|t_s|} \sum_{s \in t_s} \delta_{s}$. Based on the above sampling idea, we sample $q$ target sentence semantics from $\mathcal{P}_{t_s}$ instead of using the entire set, denoted as $\{y^{tar}_{i} \vert 1 \leq i \leq q\}$. Next, we also obtain $q$ samples from Gaussian Distribution $\mathcal{N}(\mu_{f_3}, \Sigma_{f_3})$ using reparameterization trick \cite{kingma2013reptick}, denoted as $\{\xi_{i} \sim  \mathcal{N}(\mu_{f_3}, \Sigma_{f_3}) \vert 1 \leq i \leq q\}$. Following \cite{yang2024improving}, we feed each $\xi_{i}$ into FNN to get $q$  prediction information, denoted as $\{y^{pre}_{i}=\mathrm{FNN}^{f_{3}}(\xi_{i}) \vert 1 \leq i \leq q\}$. To this end, we can minimize the Wasserstein distance between $y^{tar}$ and $y^{pre}$ to achieve semantic alignment of $Z_{f_3}$ and $t_s$. That is:
\begin{equation}
    \mathcal{L}_{t_s} = \mathcal{W} (y^{pre}, y^{tar}).
\label{eq:sentence_w2_loss}
\end{equation}
Finally, we apply the Sinkhorn-Knopp algorithm \cite{cuturi2013lightspeed} to approximate the solution, whose complexities are $O(q^2)$. 

\begin{theorem}
    Let \( \mu \in \mathcal{P}_2(\mathbb{R}^d) \) be absolutely continuous with respect to the Lebesgue measure with Radon–Nikodym density \( \rho(x) \). Let \( \nu = \sum_{i=1}^n \nu_i \delta_{y_i} \) for some \( \{y_j\}_{j=1}^n \subset \mathbb{R}^d \), \( \nu_j \geq 0 \) and \( \sum_{j=1}^n \nu_j = 1 \), where $\delta$ is Dirac delta function. Then, for any $\epsilon > 0$, there exists a fully connected deep neural network \( u (\cdot): \mathbb{R}^d \to \mathbb{R} \) with sufficiently large width and depth (depending on $\epsilon > 0$) such that the Wasserstein distance between $\nabla u(\mu)$ and $\nu$ is less than $\epsilon$, where $\nabla u(\cdot)$ is gradient of $u(\cdot)$.
\label{tr:FNN}
\end{theorem}

\subsection{Training Objective}
Following the above process, we can also calculate the loss $\mathcal{L}_{t_w}$ (\ie, $Z_{f_1}$ and words $t_w$) and $\mathcal{L}_{t_p}$ (\ie, $Z_{f_2}$ and phrases $t_p$). We use three hyperparameters (\ie, $\alpha$, $\beta$, and $\gamma$) to control three semantic alignment losses, respectively. Finally, the overall objective function is:
\begin{equation}
    \mathcal{L} = \mathcal{L}_{vq} + \alpha \mathcal{L}_{t_w} + \beta \mathcal{L}_{t_p} + \gamma \mathcal{L}_{t_s}.
\label{eq:losses}
\end{equation}

\section{Experiments}
\subsection{Experimental Settings}
\textbf{Benchmarks}. Our method is model-agnostic, allowing it to be applied to different VQ-based architectures. Then, we choose VQ-GAN \cite{esser2021taming} and CVQ \cite{Zheng2023CVQ} as our backbone networks. Additionally, our method incorporates pre-trained text semantics, to ensure a fair comparison, we use recent works LG-VQ \cite{liang2024lgvq}\textsubscript{NeurIPS'2024} and VQCT \cite{zhang2024vqct}\textsubscript{CVPR'2024} as baselines, both of which also leverage language materials and the pre-trained language model. We evaluate our method on three public datasets for image reconstruction, including CelebA-HQ~\cite{liu2015cele}, CUB-200~\cite{wah2011cub}, and MS-COCO~\cite{lin2014coco}. We also evaluate our method on various downstream tasks based on the learned codebook, including text-to-image, semantic synthesis, unconditional image generation, and image completion on CelebA-HQ \cite{liu2015cele}, and image captioning on CUB-200~\cite{wah2011cub}, and visual grounding on refcoco dataset~\cite{yu2016refcoco}, and visual question answering (VQA) on COCO-QA dataset~\cite{ren2015cocoQA}.\\
\textbf{Evaluation Metrics}. Following previous works \cite{esser2021taming,Zheng2023CVQ,Zhang2023Regularized,zhang2024vqct}, we adopt the standard Fréchet Inception Distance (FID) \cite{heusel2017fid} to evaluate the quality of image reconstruction and generation. For other downstream tasks, such as visual grounding, image captioning, and VQA, we follow the evaluation strategy outlined in \cite{liang2024lgvq}.\\
\textbf{Implementation Details}. Following VQ-GAN and LG-VQ \cite{esser2021taming,liang2024lgvq}, all images are reshaped 256 $\times$ 256 for reconstruction and generation. We set the down-sampling factor $f$ to 16 and the codebook size $K$ to 1024, with a batch size of 6. The architecture of TA-VQ exactly follows VQ-GAN except for the proposed hierarchical encoder and semantic alignment module. 
All FNNs are 2-layer multilayer perception (MLP) with ReLU activation function. The three key hyperparameters ($\alpha$, $\beta$, and $\gamma$) are uniformly set to 0.001 across all datasets. For further details regarding the generated text, please refer to the supplementary materials. 
\subsection{Discussion of Results}
\cref{tab:reconexp} compares our method and the state-of-the-art VQ-based method in image reconstruction performance. From the results, we can obtain several key conclusions: 1) Compared with the backbone network (\ie, VQ-GAN, and CVQ), our method can learn more robust codebook representation by introducing pre-trained text semantics to significantly improve reconstruction performance, indicating the effectiveness of our method; 2) Our method surpasses LG-VQ across all datasets, largely due to its ability to integrate long text with richer semantic knowledge and effectively encode them at different granularities to learn more robust codebook learning; 3) Compared to VQCT, our method can achieve lower reconstruction errors when using the CVQ backbone, which suggests the effectiveness and generality of method. Notably, our method achieves superior reconstruction and significantly enhances codebook performance across various downstream tasks, suggesting our method's effectiveness. Further qualitative comparisons are provided in the supplementary materials. 
\begin{table}[t]
\footnotesize
\resizebox{\linewidth}{!}{
\renewcommand\tabcolsep{1.25pt}
\centering 
\begin{tabular}{lccccc}
\bottomrule
\multirow{2}{*}{Models} & Codebook & \#Tokens & CelebA-HQ       & CUB-200         & MS-COCO         \\
                        &   Size      &          & FID$\downarrow$ & FID$\downarrow$ & FID$\downarrow$ \\ \hline
VQCT \cite{zhang2024vqct}      & 6207          & 512      & 5.02            & 4.52            & 9.82            \\ \hline
VQ-GAN \cite{esser2021taming}  & 1024          & 256      & 5.66            & 5.31            & 14.45           \\
LG-VQ \cite{liang2024lgvq}     & 1024          & 256      & 5.34            & 4.74            & 10.72           \\ \rowcolor{gray!40}
TA-VQ (Ours)                  & 1024          & 256      & 5.03            & 4.60            & 10.32           \\ \toprule
CVQ \cite{Zheng2023CVQ} & 1024          & 256      & 5.19            & 4.64            & 9.94            \\
LG-CVQ \cite{liang2024lgvq}  & 1024          & 256      & 4.90            & 4.40            & 9.69            \\ \rowcolor{gray!40}
TA-CVQ (Ours)               & 1024          & 256      & \textbf{4.71}            & \textbf{4.03}               & \textbf{9.65}               \\ \toprule
\end{tabular}
}
\caption{
Results (FID$\downarrow$) of image reconstruction on CelebA-HQ, CUB-200, and MS-COCO. The best results are highlighted in bold.
}
\label{tab:reconexp}
\vspace{-3mm}
\end{table}

\begin{table}[t]
    \centering
    \footnotesize
    \begin{tabular}{llccc}
    \bottomrule
    \multirow{2}{*}{} & \multirow{2}{*}{Setting}                                           & CUB-200 \\ %\cline{3-4} 
                      &                                                                    & FID$\downarrow$     \\ \hline
    (i)               & Baseline(VQ-GAN)                                                    & 5.51    \\
    (ii)             & + $\mathcal{L}_{s}$                                                 & 4.89    \\
    (iii)              & + $\mathcal{L}_{s}$ + $\mathcal{L}_{p}$                           & 4.69    \\
    (iv)               & + $\mathcal{L}_{s}$ + $\mathcal{L}_{w}$                           & 4.82       \\
    (v)               & + $\mathcal{L}_{w}$ + $\mathcal{L}_{p}$                           & 4.75       \\
    (vi)              & + $\mathcal{L}_{w}$ + $\mathcal{L}_{p}$ + $\mathcal{L}_{s}$       & 4.60    \\ \toprule
    \end{tabular}
    \caption{Ablation study of our three loss functions on CUB-200. ``+$\mathcal{L}_{s}$'' denotes introduces the sentence semantic alignment into the baseline model.}
    \label{tab:ablation}
    \vspace{-4mm}
\end{table}
\begin{table*}[t]
    \centering
    \footnotesize
    \setlength{\tabcolsep}{0pt}
    \renewcommand{\arraystretch}{1.0}
    \begin{minipage}[t]{0.20\textwidth}
        \centering
        \begin{tabular}{lc}
        \bottomrule
        Setting               & \multicolumn{1}{l}{FID$\downarrow$} \\ \hline
        w/o multi-granularity & 4.91                                 \\
        TA-VQ                 & 4.60                                 \\ \toprule
        \end{tabular}
        \caption{Ablation study of multi-granularity text encoding on CUB-200. ``w/o'' means without.}
        \label{tab:multigranularity}
    \end{minipage}
    \hfill
    \begin{minipage}[t]{0.20\textwidth}
        \centering
        \begin{tabular}{lc}
        \bottomrule
        \multicolumn{1}{l}{Setting} & \multicolumn{1}{l}{Similarity$\uparrow$} \\ \hline
        LG-VQ \cite{liang2024lgvq}                       & 0.1021                                    \\
        TA-VQ                       & 0.1444                                    \\ \toprule
        \end{tabular}
        \caption{Ablation study of cosine similarity evaluation between codebook and text on CUB-200.}
        \label{tab:similarity}
    \end{minipage}
    \hfill
    \begin{minipage}[t]{0.2\textwidth}
        \centering
        \begin{tabular}{lc}
        \bottomrule
        \multirow{2}{*}{Setting} & \#Train time  \\
                                    &  sec/epoch$\downarrow$              \\ \hline
        w/o sampling                & 2,065                    \\
        TA-VQ                       & 1,930                    \\ \toprule
        \end{tabular}
        \caption{Ablation study of computational overhead without and with sampling strategy.}
        \label{tab:compute_overhead}
    \end{minipage}
    \hfill
    \begin{minipage}[t]{0.2\textwidth}
        \centering
        \begin{tabular}{lc}
        \bottomrule
        \multirow{2}{*}{Model} & \multicolumn{1}{c}{Image Completion}                               \\
                                     & FID$\downarrow$                                             \\ \hline
        
        VQ-GAN                        & 9.02   \\
        LG-VQ                    & 8.14   \\ 
        TA-VQ                   & \textbf{8.04} \\ \toprule
        \end{tabular}
        \caption{Result (FID$\downarrow$) of image completion on CelebA-HQ.}
        \label{tab:imagecompletion}
    \end{minipage}
    \vspace{-4mm}
\end{table*}
\subsection{Ablation Study}
\textbf{Are our semantic alignments of words, phrases, and sentences both effective?} 
We conduct extensive ablation studies to evaluate the impact of three semantic losses, as summarized in~\cref{tab:ablation}. We use VQ-GAN as the baseline model, without incorporating any additional semantic losses. From the results, several key conclusions can be drawn: First, all three semantic losses are essential for enhancing reconstruction performance. Second, the results of (i) to (iv) show that sentence-level semantics ($\mathcal{L}_{s}$) contribute the most to performance improvement. This can be attributed to the richer, high-level semantic information sentences provide, which supports more robust codebook learning. Furthermore, because the alignment takes place at the level of codes ($Z_{f_{3}}$) used for reconstruction, the semantic knowledge can directly be utilized to improve reconstruction quality. Finally, comparing (vi) with (i) to (v), we observe that incorporating all three losses results in the best overall performance, demonstrating the effectiveness of our method.\\
\textbf{Is our multi-granularity text encoding method effective?} To answer the question, we analyze the impact of not having multi-granularity text encoding, \ie using the entire long text semantics directly aligned with the image codes. The results are shown in~\cref{tab:multigranularity}, we can observe that considering multi-granularity text encoding can achieve better reconstruction, which is reasonable because such encoding manner can effectively retain all key semantic knowledge. This shows the effectiveness of our method.  \\
\textbf{Can our codebook be effectively aligned with the text?} In~\cref{tab:similarity}, we measure the cosine similarity between the codebook and text on CUB-200 all test data. From the results, we can see that TA-VQ's codebook is more similar to the text than LG-VQ, indicating that our method can learn better text-aligned codebook. \\
\textbf{Can our sampling-based alignment strategy reduce the computational overhead?} To answer the question, we conduct an experiment without and with sampling alignment strategy with a batch size equal to 1 on the CUB-200 dataset. The experiments are conducted on Debian GNU/Linux 12 with Intel(R) Core(TM) i9-10900X CPU, a GTX 4090, and 128G memory. We calculate the seconds required for both methods to train one epoch, the results are shown in~\cref{tab:compute_overhead}, which suggests our design of sampling-based alignment strategy can effectively reduce the computational overhead. 
\begin{figure}[t]
    \centering
    \includegraphics[width=1\linewidth]{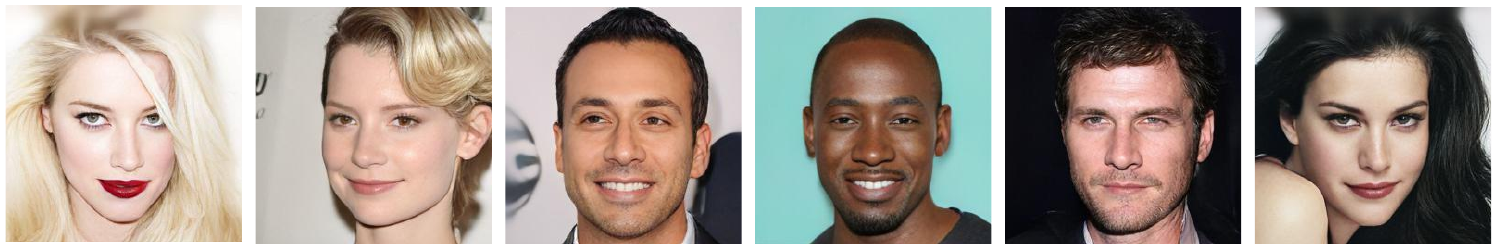}
    \caption{Examples of unconditional generation on CelebA-HQ. More examples are provided in supplementary materials.}
    \label{fig:uncond}
    \vspace{-2mm}
\end{figure}
\begin{figure}[t]
    \centering
    \includegraphics[width=1\linewidth]{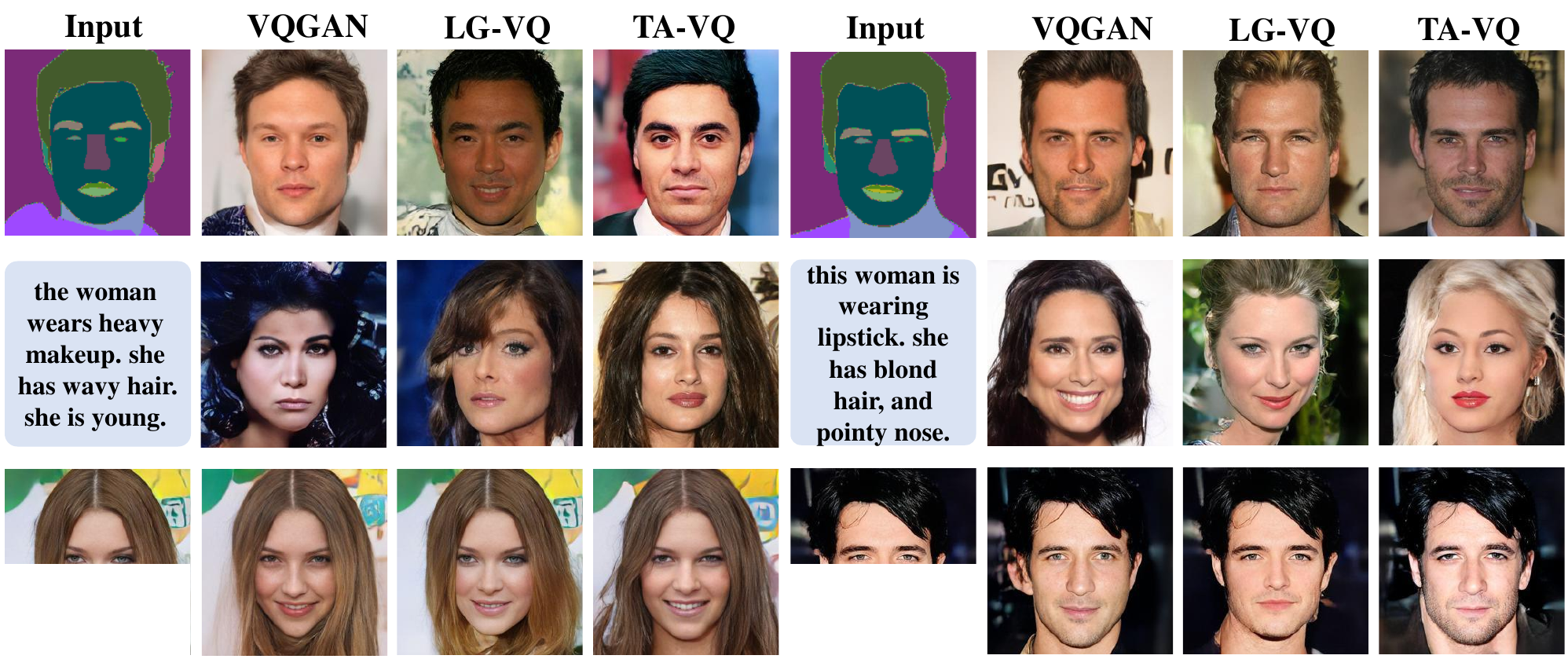}
    \caption{Examples of semantic synthesis (row 1), text-to-image (row 2), and image completion (row 3). More examples are provided in supplementary materials.}
    \label{fig:conditional}
    \vspace{-3mm}
\end{figure}
\begin{table*}[t]
    \centering
    \footnotesize
    \begin{minipage}[t]{0.30\textwidth}
        \centering
        \begin{tabular}{lc}
        \bottomrule
            \multirow{2}{*}{Model} & \multicolumn{1}{c}{Unconditional Generation}                               \\
                                     & FID$\downarrow$                                             \\ \hline
        DC-VAE \cite{pidhorskyi2020adversarial}                        & 15.8    \\
        VQ-GAN  \cite{esser2021taming}                      & 10.2   \\
        LG-VQ \cite{liang2024lgvq}                    & 9.1   \\ 
        TA-VQ               & \textbf{8.8}   \\ \toprule
        \end{tabular}
        \caption{Result (FID$\downarrow$) of unconditional image generation on CelebA-HQ.}
        \label{tab:uncond_generation}
    \end{minipage}
    \hfill
    \begin{minipage}[t]{0.30\textwidth}
        \centering
        \begin{tabular}{lc}
        \bottomrule
        \multirow{2}{*}{Model} & \multicolumn{1}{c}{Semantic Synthesis}                               \\
                                     & FID$\downarrow$                                             \\ \hline
        VQCT \cite{zhang2024vqct}                     & 14.47   \\
        VQ-GAN \cite{esser2021taming}                        & 11.53   \\
        LG-VQ \cite{liang2024lgvq}                    & 11.46   \\ 
        TA-VQ                   & \textbf{10.74} \\ \toprule
        \end{tabular}
        \caption{Result (FID$\downarrow$) of semantic synthesis on CelebA-HQ.}
        \label{tab:semantic_syn}
    \end{minipage}
    \hfill
    \begin{minipage}[t]{0.30\textwidth}
        \centering
        \begin{tabular}{lc}
        \bottomrule
        \multirow{2}{*}{Model}    & Visual Grounding        \\
                                  & Accuracy(0.5)$\uparrow$ \\ \hline
        VQ-GAN \cite{esser2021taming}     & 9.14                    \\
        VQCT \cite{zhang2024vqct} & 9.46                    \\
        LG-VQ \cite{liang2024lgvq} & 9.62                    \\ 
        TA-VQ                     & \textbf{10.17}                   \\ \toprule
        \end{tabular}
        \caption{Result of visual grounding on refcoco dataset \cite{yu2016refcoco}.} %using MS-COCO's codebook.}
        \label{tab:visual_grounding}
    \end{minipage}
% \vspace{-1mm}
\end{table*}
\begin{table*}
    \centering
    \footnotesize
    \setlength{\tabcolsep}{0pt} % Remove extra padding
    \renewcommand{\arraystretch}{1.1} % Uniform row height
    \begin{minipage}[t]{0.25\textwidth}
        \centering
        \begin{tabular}{lcc}
        \bottomrule
        \multirow{2}{*}{Model} & \multicolumn{1}{c}{Text-to-Image}                               \\
                                 & FID$\downarrow$                                             \\ \hline
        Corgi \cite{zhou2023corgi}               & 19.74   \\ 
        LAFITE \cite{zhou2022lafite}              & 12.54   \\  
        VQ-GAN \cite{esser2021taming}                     & 15.29   \\
        CVQ  \cite{Zheng2023CVQ}                       & 13.23   \\ 
        LG-VQ \cite{liang2024lgvq}               & 12.61   \\   
        TA-VQ          & \textbf{11.97}  \\  \toprule
        \end{tabular}
        \vspace{-2mm}
        \caption{Results (FID$\downarrow$) of text-to-image on CelebA-HQ. }
        \label{tab:t2image}
    \end{minipage}
    \hfill
    \begin{minipage}[t]{0.33\textwidth}
        \centering
        \begin{tabular}{lcccc}
        \bottomrule
        \multirow{2}{*}{Model}                                            & \multicolumn{3}{c}{Image Captioning}       \\
                                                         & BLEU4$\uparrow$ & ROUGE-L$\uparrow$ & METEOR$\uparrow$ & CIDEr-D$\uparrow$ \\ \hline
        
        VQ-GAN \cite{esser2021taming}                      & 1.29  & 33.40   & 24.47  & 93.62   \\
        VQCT \cite{zhang2024vqct}                       & 1.38  & 26.50   & 24.63  & 98.22 \\
        LG-VQ \cite{liang2024lgvq}                      & 1.69  & 34.73   & 25.78  & 102.77   \\ 
        TA-VQ                 & \textbf{1.90}  & \textbf{35.50}   & \textbf{27.61}  & \textbf{109.42}   \\ 
        \toprule
        \end{tabular}
        \caption{Results of image captioning on CUB-200 datasets.}
        \label{tab:imagecaption}
    \end{minipage}
    \hfill
    \begin{minipage}[t]{0.30\textwidth}
        \centering
        \begin{tabular}{lcc}
        \bottomrule
        \multirow{2}{*}{Setting} & \multicolumn{2}{c}{VQA}                               \\
                                 & Accuracy$\uparrow$                       & WUPS$\uparrow$\cite{wu1994wups}                      \\ \hline
        VQCT \cite{zhang2024vqct} & 40.42                     & 82.06                     \\
        VQ-GAN \cite{esser2021taming}         & 37.82                     & 83.22                     \\
        LG-VQ \cite{liang2024lgvq}       & 40.97           & 83.56 \\ 
        TA-VQ       & \textbf{41.56}        & \textbf{83.77}  \\\toprule
        \end{tabular}
        \caption{Results of VQA on COCO-QA~\cite{ren2015cocoQA} dataset.}% using MS-COCO's codebook.}
        \label{tab:vqa}
    \end{minipage}
% \vspace{-1mm}
\end{table*}
\subsection{Application} \label{sec:application}
To further assess the effectiveness of our method, we apply the codebook learned by our method to various downstream tasks, including image generation, visual grounding, and visual text reasoning. We select the VQ-GAN as the backbone network, named TA-VQ.
\subsubsection{Image Generation}
Following \cite{esser2021taming,liang2024lgvq,rombach2022high,gu2022vqdiffusion}, we conduct image generation downstream tasks, \ie image completion, unconditional and conditional image generation, to verify the effectiveness of our method. \\
%For a more detailed experimental setting, please refer to the supplementary material. \\
\textbf{Image Completion and Unconditional Generation }. We follow the default setting of VQ-GAN-transformer \cite{esser2021taming} and conduct image completion and unconditional image generation on CelebA-HQ. The results are shown in~\cref{tab:imagecompletion} and~\cref{tab:uncond_generation}, respectively.  From the results, our method can achieve superior performance, which indicates that our approach can more effectively leverage text semantics to learn a robust codebook representation, suggesting the effectiveness of our method. In comparison to LG-VQ, our method can further improve performance, which confirms our motivation for using longer text. Examples of Unconditional generation are shown in ~\cref{fig:uncond}. We also provide a qualitative comparison of image completion in~\cref{fig:conditional} row 3.
\textbf{Conditional Generation}. We further evaluate the performance of our codebook on two tasks: semantic synthesis \cite{esser2021taming} and text-to-image generation \cite{gu2022vqdiffusion}. For the semantic synthesis task, we follow the default settings of VQ-GAN-transformer, with results provided in~\cref{tab:semantic_syn}. TA-VQ achieves the best performance compared to other models. This can be attributed to our multi-granularity encoding design, which fully leverages the rich semantic information in long texts for more robust codebook learning. For the text-to-image task, we use VQ-diffusion \cite{gu2022vqdiffusion} as the backbone network, with results shown in~\cref{tab:t2image}. TA-VQ achieves a lower FID score, primarily due to our multi-hierarchical semantic alignment module, which enables a more comprehensive understanding of text conditions. This supports the effectiveness of our method. Qualitative comparison of conditional generation can be found in~\cref{fig:conditional} row 1 and row 2.
\subsubsection{Visual Grounding}
\begin{figure}
    \centering
    \includegraphics[width=1\linewidth]{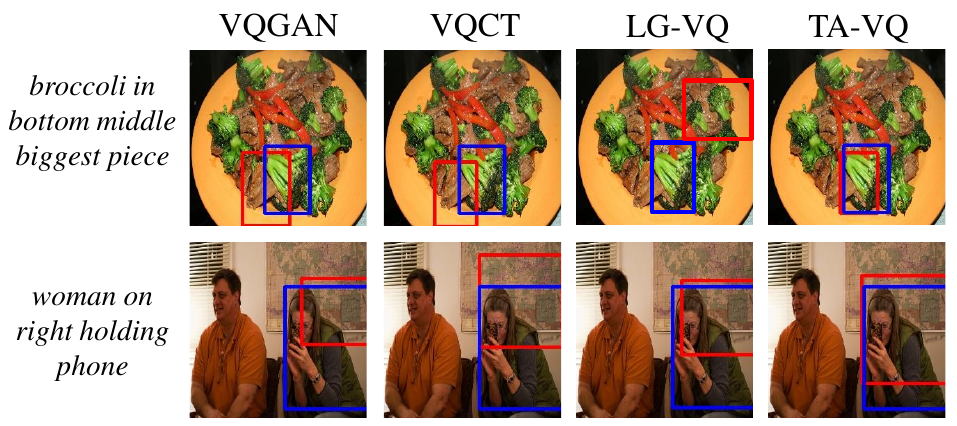}
    \caption{Visualizations for visual grounding. The blue boxes are the ground-truth, red boxes are the model predictions. More Examples are provided in supplementary materials.}
    \label{fig:vg}
    % \vspace{-3mm}
\end{figure}

We conduct a visual grounding task on refcoco dataset \cite{yu2016refcoco} utilizing MS-COCO's codebook. The evaluation metric follows prior work \cite{deng2021transvg,liang2024lgvq}, where a prediction is considered correct if the IoU between the ground truth box and the predicted bounding box exceeds 0.5. We select VQ-GAN, VQCT, and LG-VQ as baseline. The results are shown in~\cref{tab:visual_grounding}, TA-VQ achieves the best performance compared with baseline models. We provide a qualitative comparison in~\cref{fig:vg}, which shows TA-VQ achieves more accurate predictions. This confirms the advantage of introducing longer text to learn an improved text-aligned codebook. 
\begin{figure}
    \centering
    \includegraphics[width=1\linewidth]{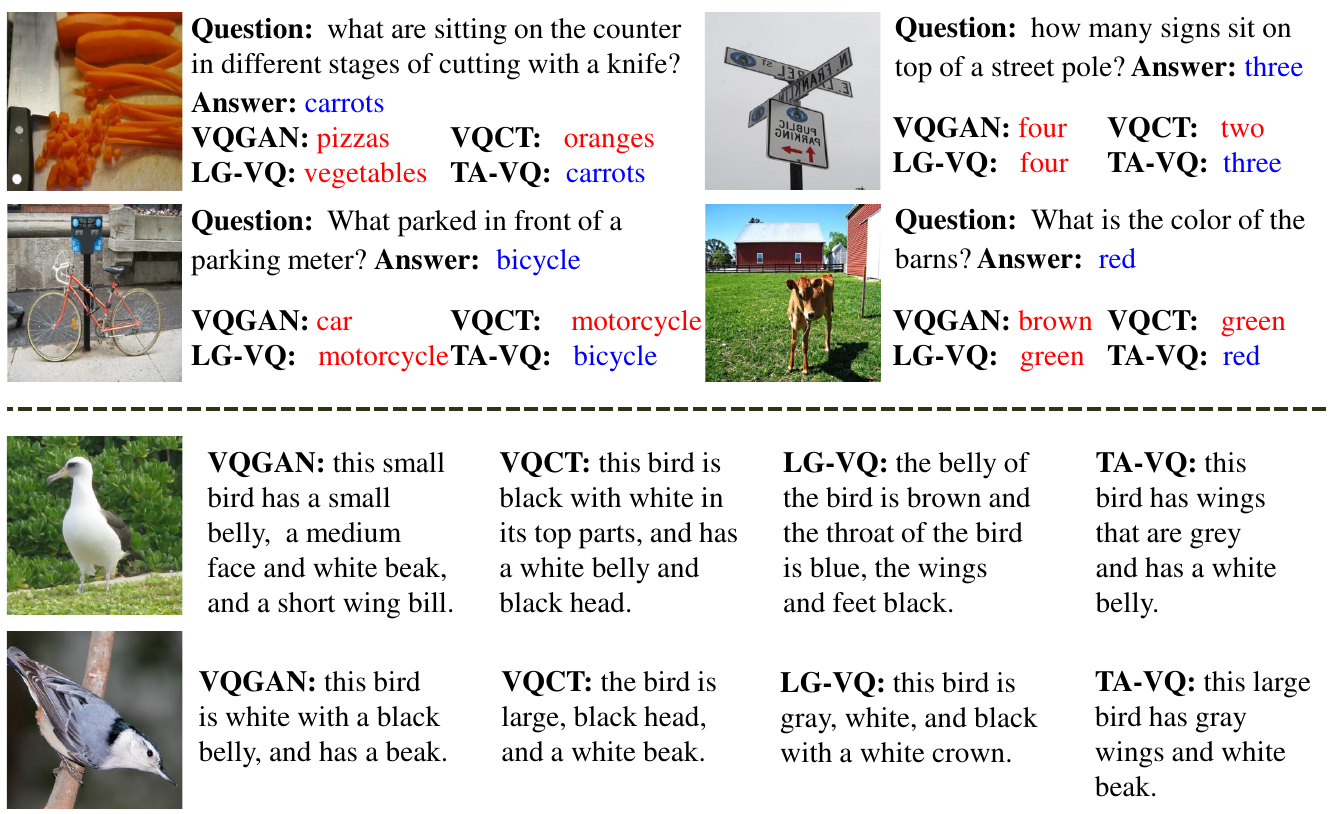}
    \caption{Visualizations for VQA, and image captioning. More Examples are provided in supplementary materials.}
    \label{fig:vrt}
    \vspace{-3mm}
\end{figure}
\subsubsection{Visual Text Reasoning}
To fully verify the effectiveness of the learned codebook, following \cite{mokady2021clipcap,ding2021cogview,liang2024lgvq}, we conduct two visual text reasoning tasks, \ie, image captioning on CUB-200, and visual question answering (VQA) on COCO-QA dataset \cite{ren2015cocoQA} using MS-COCO's codebook. Following \cite{liang2024lgvq}, we use a pre-trained language model (\eg, GPT2 \cite{radford2019gpt}) as the backbone network. We select VQ-GAN, VQCT, and LG-VQ as baseline. The results are shown in~\cref{tab:imagecaption} for image captioning and~\cref{tab:vqa} for VQA. From the results, we can see that our method outperforms the baseline models on all tasks. This is likely because the design of multi-granularity text encoding and sampling-based alignment enhances the model’s ability to achieve fine-grained codebook-text alignment, which improves the performance of cross-modal tasks. Finally, we provide some examples in~\cref{fig:vrt}, which shows that our method can accurately generate text with consistent semantics with the image in the image captioning, and can accurately answer the question in the VQA. This confirms the effectiveness of our approach.

\section{Conclusions}
This paper proposes a novel text-augmented codebook learning framework, named TA-VQ, which utilizes VLMs to generate longer text for improved text-aligned codebook learning. In particular, we propose to encode the long text semantics from three granularities, \ie, word, phrase, and sentence. Moreover, we also design a novel sampling-based alignment strategy for achieving fine-grained codebook text alignment, which does not introduce much computational overhead. Both quantitative and qualitative experiments show that our approach achieves superior reconstruction and substantially improves codebook performance across various downstream tasks, demonstrating the effectiveness of our method.

\newpage

\section*{Acknowledgement} 
This work was supported by the Guangdong Basic and Applied Basic Research Foundation under Grant No. 2025A1515011674,  the Shenzhen Peacock Program under Grant No. ZX20230597, NSFC under Grant No. 62272130 and Grant No. 62376072. It was also supported by the Major Key Project of Peng Cheng Laboratory. 

{
    \small
    \bibliographystyle{ieeenat_fullname}
    \bibliography{main}
}
\newpage
% % WARNING: do not forget to delete the supplementary pages from your submission 
\clearpage
\setcounter{page}{1}
\maketitlesupplementary
% \subsection{Application Datasets}

\section{Experiment Details} \label{App:expdetails}
\textbf{Image Generation}: For semantic image synthesis, unconditional generation, and image completion, we follow the default setting of VQ-GAN-Transformer\footnote{https://github.com/CompVis/taming-transformers}. We use a 16-layer transformer as the backbone network with the head of 16 and dimension of 1024. More settings can be found in~\cite{esser2021taming}.
For the text-to-image task, we follow the default setting of VQ-Diffusion\footnote{https://github.com/microsoft/VQ-Diffusion}. We use a 19-layer transformer as the backbone network with the head of 16 and dimension of 1024. The diffusion step is 100. More settings can be found in~\cite{gu2022vqdiffusion}. \\
\textbf{Visual Grounding}: The RefCOCO dataset \cite{yu2016refcoco} comprises 19,994 images with 50,000 referred objects, each associated with multiple referring expressions, totaling 142,210 expressions. Following \cite{liang2024lgvq}, we adopt the RefCOCOgumd \cite{nagaraja2016refcocoumc} protocol to partition the dataset, resulting in 42,404 expressions for training, 3,811 for validation, and 3,785 for testing.
We follow the default setting of LG-VQ~\cite{liang2024lgvq}, we first apply the VQ model to quantize the image into discrete token representations, which are then processed by a learnable adapter network (\eg, a 2-layer MLP). We use a pre-trained CLIP model \cite{Alec2021clip} to encode the object descriptions, and then concatenate the image token representations with the text embeddings. These concatenated features are fed into a trainable transformer, and the last hidden embeddings are used to predict the object box.
We use a 3-layer transformer as the backbone network with the head of 16 and dimension of 512. The learnable adapter network is 2-layer MLP with ReLU activation function. The prediction network of the object box is 3-layer MLP with ReLU activation function. \\
\textbf{Visual Text Reasoning}: The COCO-QA dataset~\cite{ren2015cocoQA} is automatically generated from captions in the Microsoft COCO dataset~\cite{lin2014coco} and contains 78,736 training questions and 38,948 test questions, based on 8,000 and 4,000 images, respectively. The dataset comprises four types of questions: object (70\%), number (7\%), color (17\%), and location (6\%). Each answer is a single-word.
We follow the default setting of LG-VQ~\cite{liang2024lgvq}. we first use the VQ model to quantize the image into discrete token representations, and then feed it into a learnable Adapter network (\eg, 2-layer MLP). Following this, we concatenate image token representation with the text embedding and feed them into a pre-trained language model (\ie, GPT2 \cite{radford2019gpt}). We adjust the output of the language model to adapt to different tasks.
The learnable adapter network is 2-layer MLP with ReLU activation function. For the image captioning task, we generate captions by predicting the next token in an autoregressive manner. For the VQA task, we feed the last hidden embedding into 2-layer MLP to predict the answer. 
\begin{figure}[t]
  \centering
\includegraphics[width=\columnwidth]{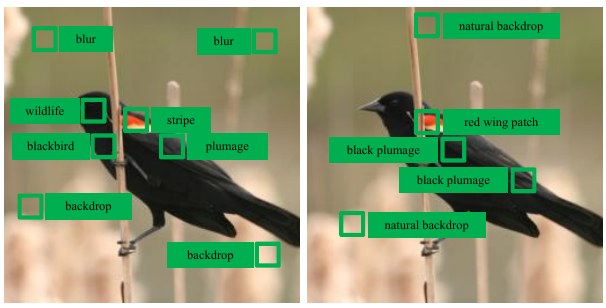}
  % \vspace{-10pt}
  \caption{Visualization of codebook-word-phrase alignment.}
  % \vspace{-18pt}
  \label{fig:align}
\end{figure}
\begin{table}[h]
\renewcommand{\thetable}{C}
        \caption{Ablation study of $f_j$ on CUB.}
	\centering
    \resizebox{\linewidth}{!}{
        \setlength{\tabcolsep}{2mm}
        {
        \begin{tabular}{l|ccccc}
            \hline
                            & [8, 16, 32] & [2, 8, 16] & [2, 4, 16] & [16, 16, 16] & [4, 8, 16] \\ \hline
            FID$\downarrow$ & 7.60    &    4.70        &   5.96     & 5.11            & \textbf{4.60}          \\ \hline
        \end{tabular}
        }
    }
	\label{tab:difflayer}
\end{table}
\section{More Ablation Study}
\textbf{Can our codebook be effectively aligned with the text?} In~\cref{fig:align}, we visualize an example to demonstrate that our method achieves satisfactory alignment. \\
\textbf{Varying the hierarchical grid features $f_j$.} We provide the impact of different $f_j$ on performance in~\cref{tab:difflayer}.

\section{Examples of Origin Caption and Long Text}
\begin{figure*}
    \centering
    \includegraphics[width=1\linewidth]{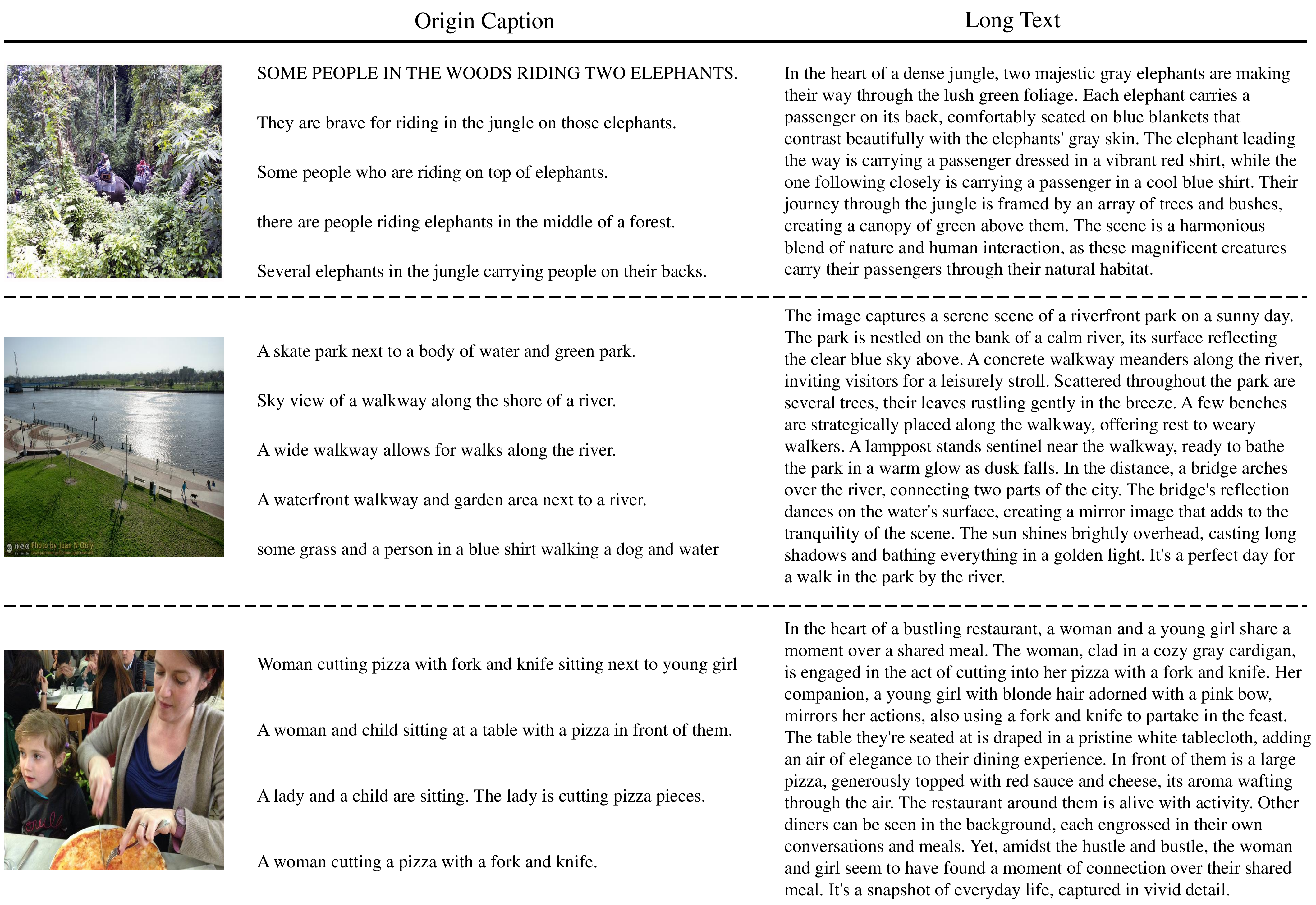}
    \caption{Examples of origin caption and long text on MS-COCO dataset.}
    \label{fig:COCO_longtext}
\end{figure*}

\begin{figure*}
    \centering
    \includegraphics[width=1\linewidth]{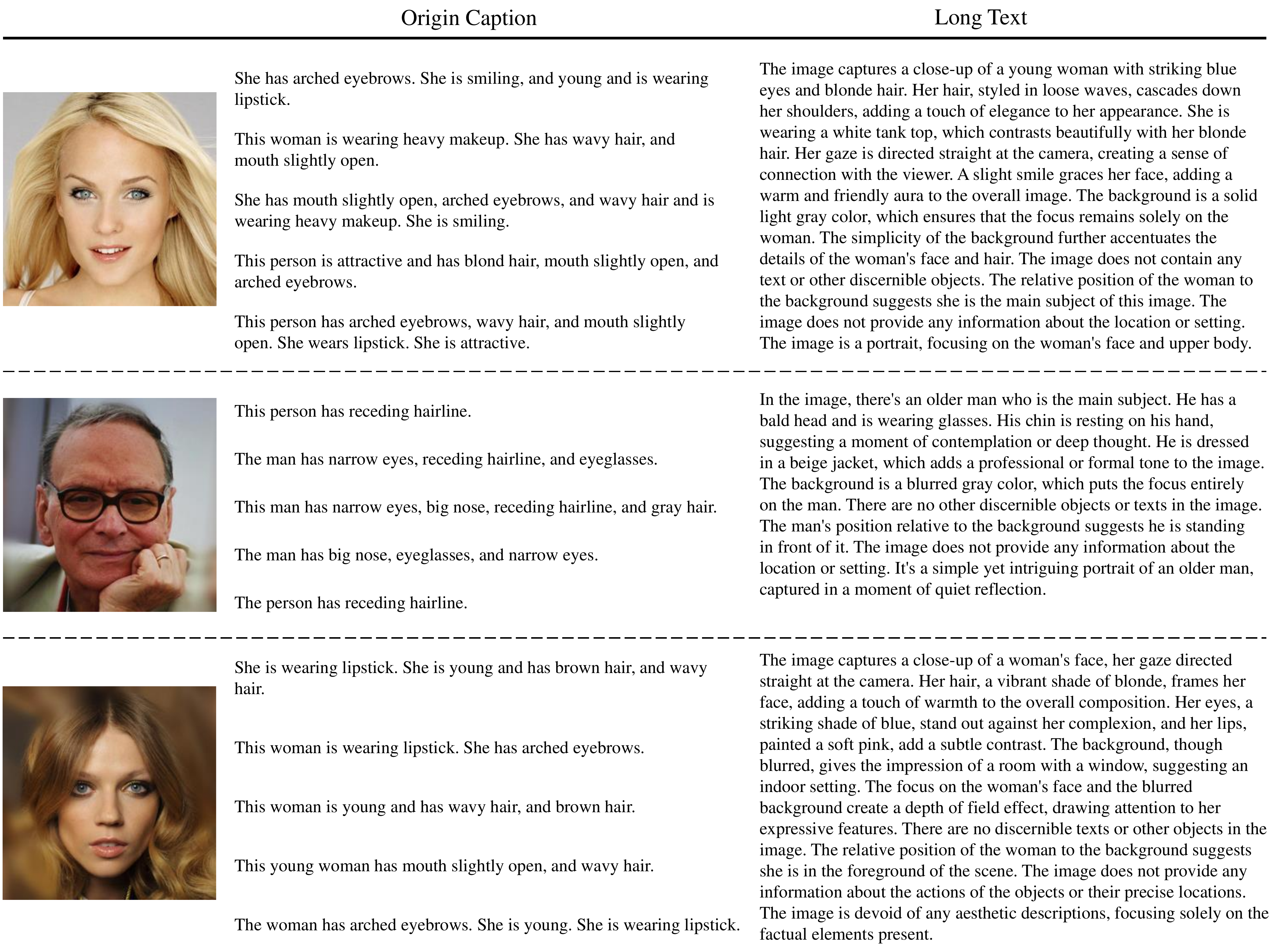}
    \caption{Examples of origin caption and long text on CelebA-HQ dataset.}
    \label{fig:cele_longtext}
\end{figure*}

\begin{figure*}
    \centering
    \includegraphics[width=1\linewidth]{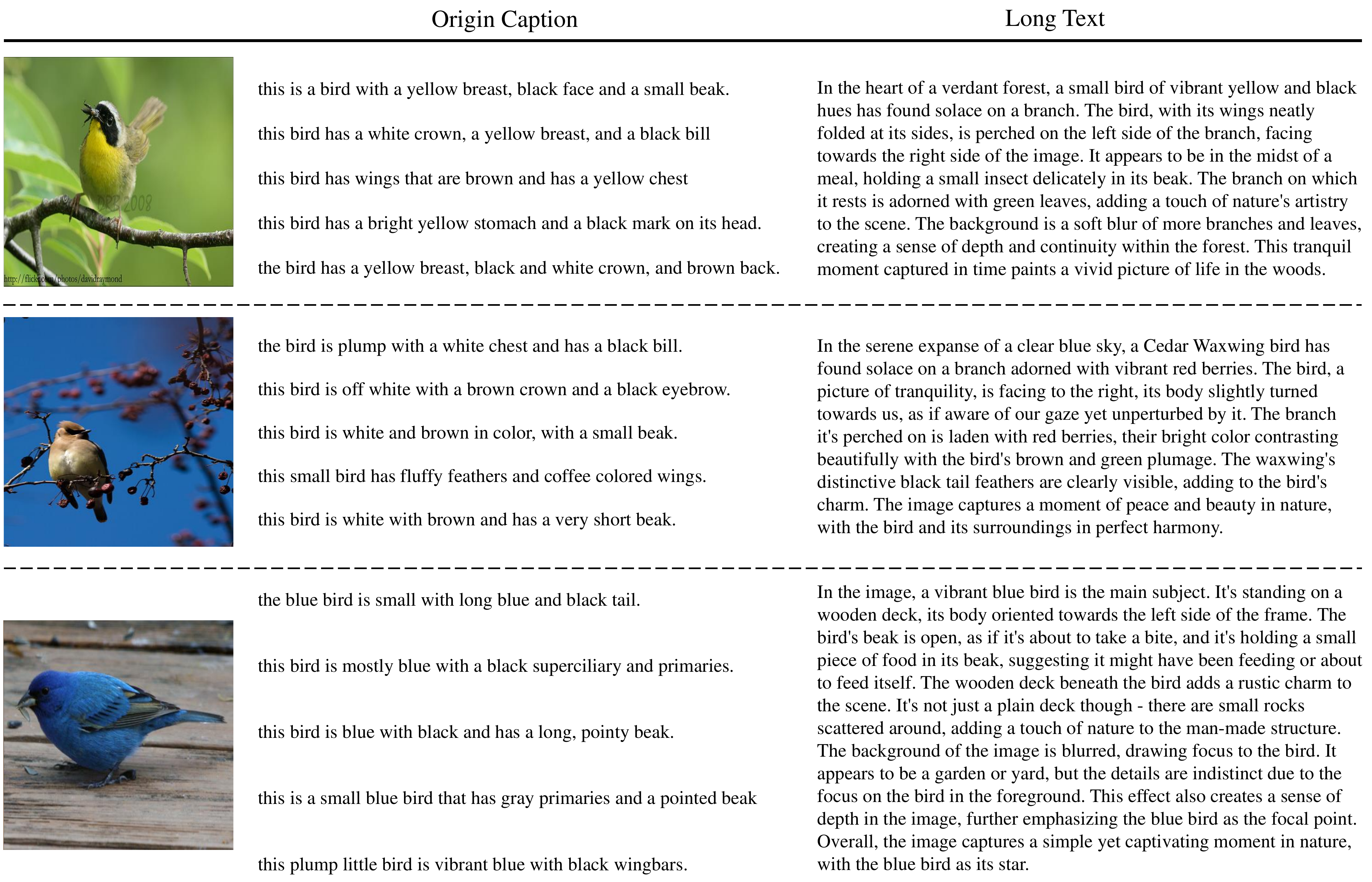}
    \caption{Examples of origin caption and long text on CUB-200 dataset.}
    \label{fig:cub_longtext}
\end{figure*}

We provide some examples of origin caption and long text in~\cref{fig:COCO_longtext} for MS-COCO dataset,  in~\cref{fig:cele_longtext} for CelebA-HQ dataset, and in~\cref{fig:cub_longtext} for CUB-200 dataset.

\section{More Examples and Qualitative Results}
\begin{figure*}
    \centering
    \includegraphics[width=1\linewidth]{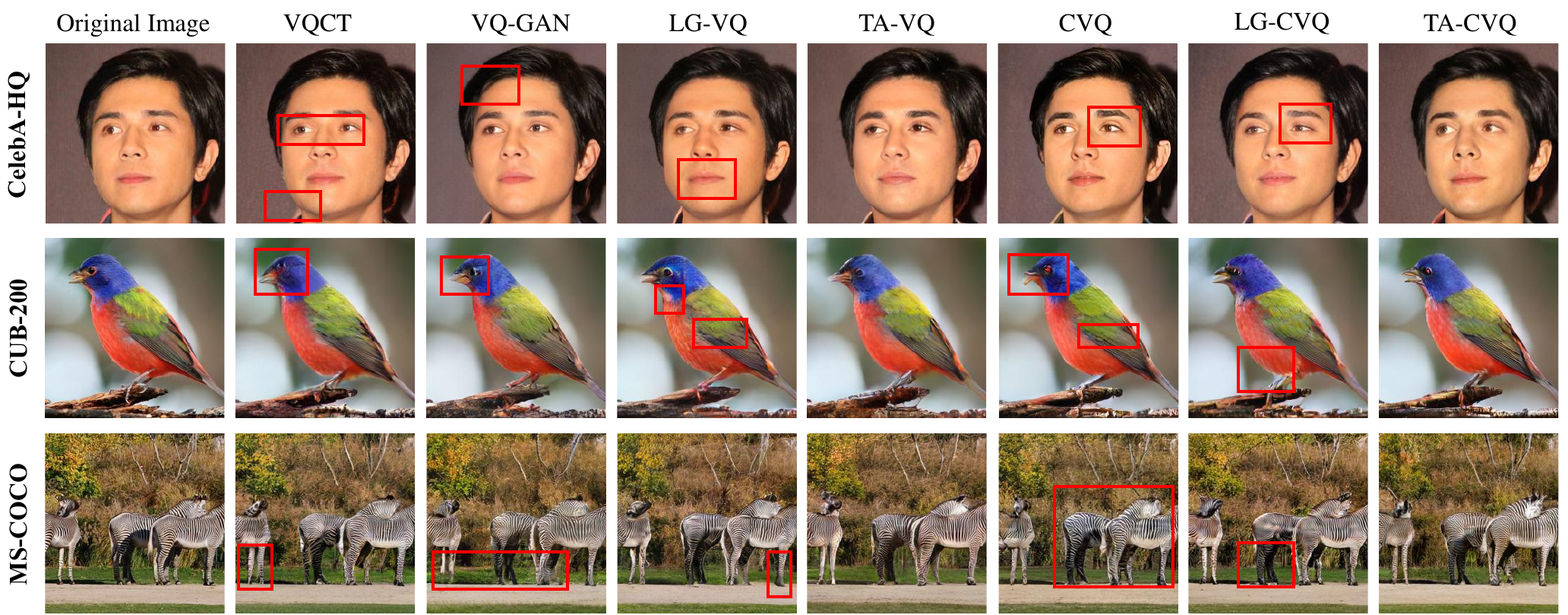}
    \caption{\textbf{Reconstructions from different models}. The red-color boxes highlight reconstruction details.}
    \label{fig:recon}
\end{figure*}

\begin{figure*}
    \centering
    \includegraphics[width=1\linewidth]{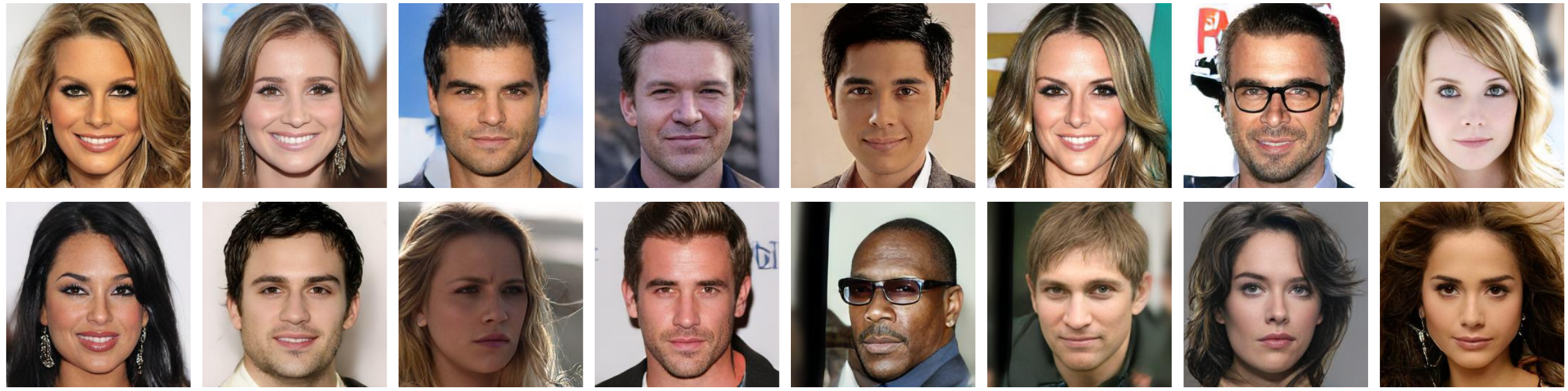}
    \caption{Visualizations for \textbf{unconditional generation} on CelebA-HQ.}
    \label{fig:uncond_append}
\end{figure*}

\begin{figure*}
    \centering
    \includegraphics[width=1\linewidth]{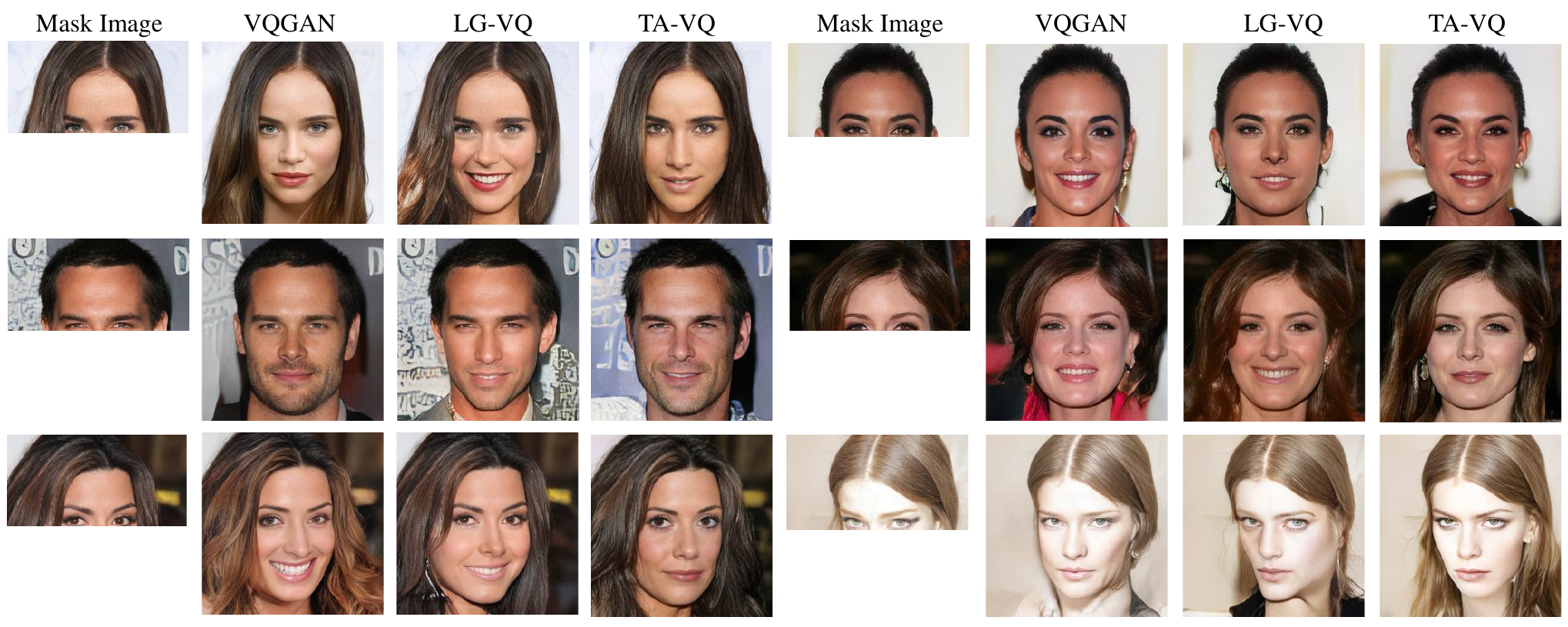}
    \caption{Visualizations for \textbf{image completion} on CelebA-HQ.}
    \label{fig:com}
\end{figure*}

\begin{figure*}
    \centering
    \includegraphics[width=1\linewidth]{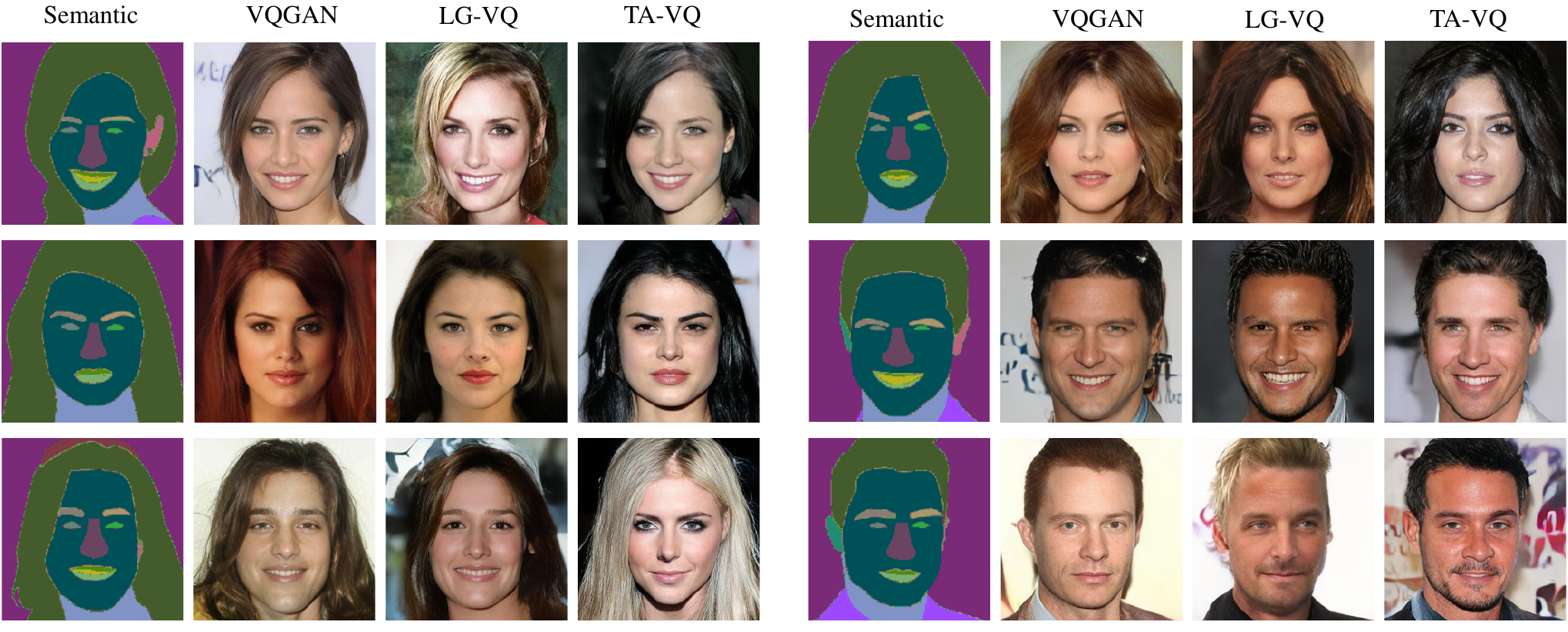}
    \caption{Visualizations for \textbf{semantic synthesis} on CelebA-HQ.}
    \label{fig:semantic}
\end{figure*}

\begin{figure*}
    \centering
    \includegraphics[width=1\linewidth]{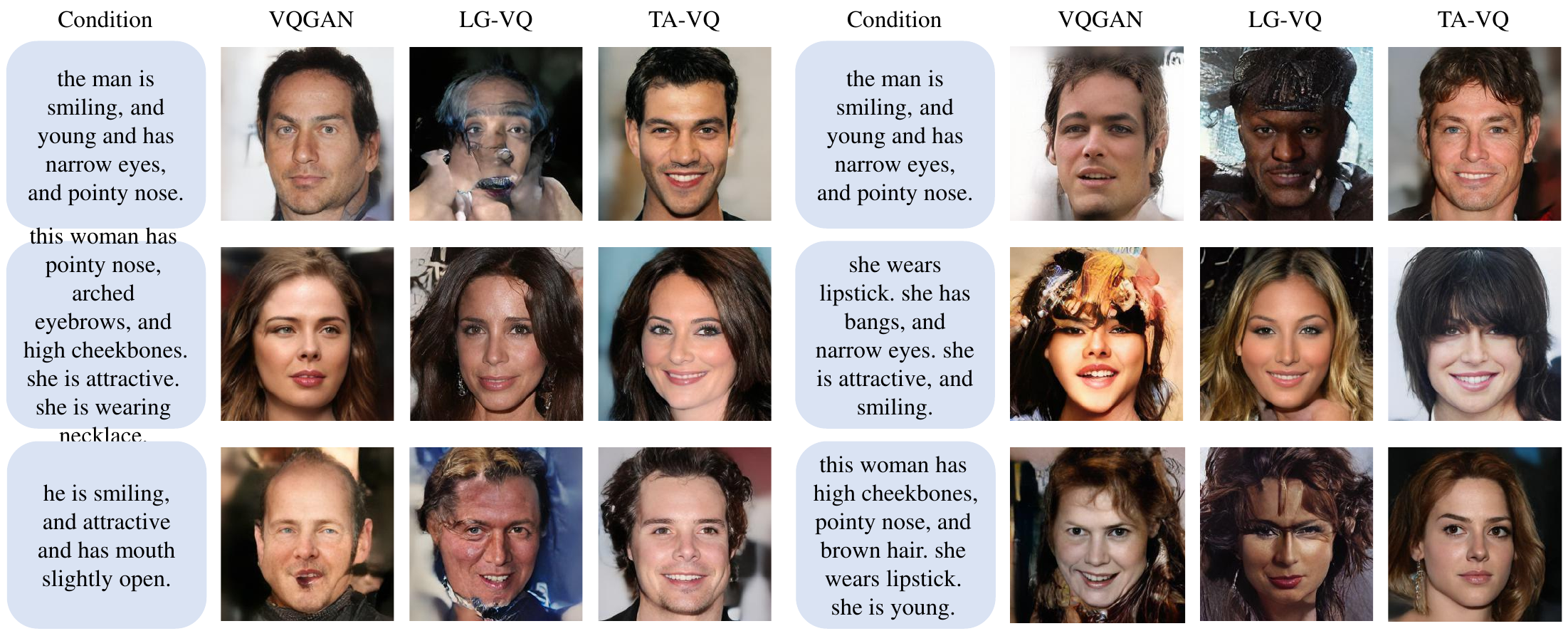}
    \caption{Visualizations for \textbf{text-to-image} on CelebA-HQ.}
    \label{fig:t2i}
\end{figure*}

\begin{figure*}
    \centering
    \includegraphics[width=1\linewidth]{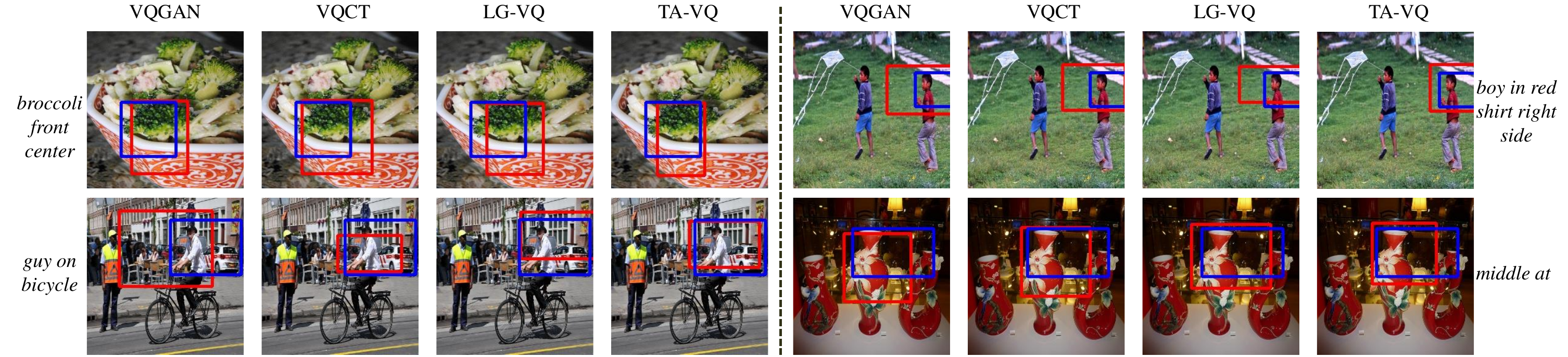}
    \caption{Visualizations for \textbf{visual grounding}. Blue boxes are the ground-truth, red boxes are the model predictions.}
    \label{fig:viusal}
\end{figure*}

\begin{figure*}
    \centering
    \includegraphics[width=1\linewidth]{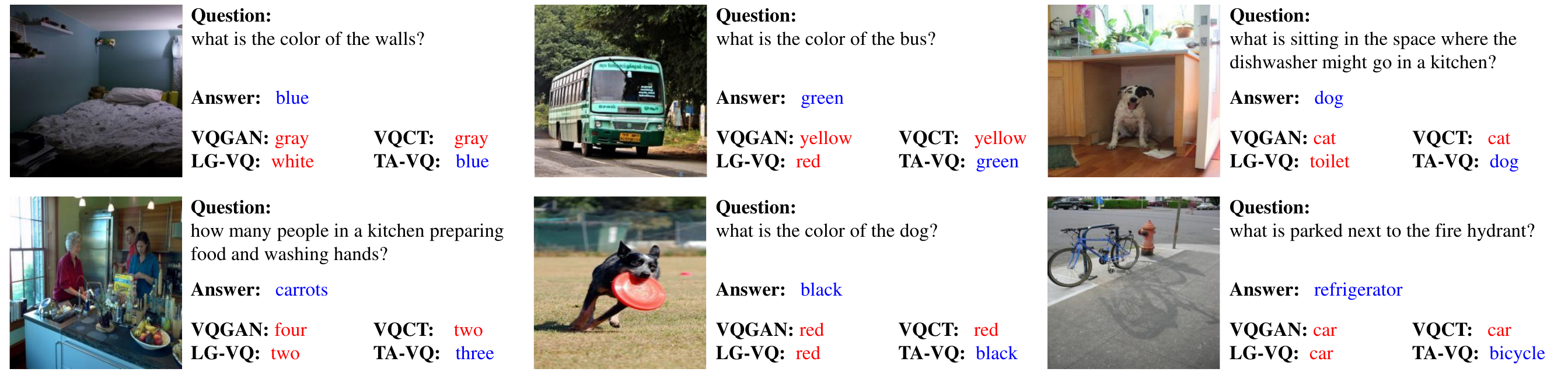}
    \caption{Visualizations for \textbf{visual question answering} (VQA).}
    \label{fig:VQA}
\end{figure*}

We provide qualitative comparison of image reconstruction in~\cref{fig:recon}, unconditional generation in \cref{fig:uncond_append}, image completion in~\cref{fig:com}, semantic image synthesis in~\cref{fig:semantic}, text-to-image in~\cref{fig:t2i}, visual grounding in~\cref{fig:viusal}, and VQA in~\cref{fig:VQA}.

% \section{Rationale}
% \label{sec:rationale}
% % 
% Having the supplementary compiled together with the main paper means that:
% % 
% \begin{itemize}
% \item The supplementary can back-reference sections of the main paper, for example, we can refer to \cref{sec:intro};
% \item The main paper can forward reference sub-sections within the supplementary explicitly (e.g. referring to a particular experiment); 
% \item When submitted to arXiv, the supplementary will already included at the end of the paper.
% \end{itemize}
% % 
% To split the supplementary pages from the main paper, you can use \href{https://support.apple.com/en-ca/guide/preview/prvw11793/mac#:~:text=Delete%20a%20page%20from%20a,or%20choose%20Edit%20%3E%20Delete).}{Preview (on macOS)}, \href{https://www.adobe.com/acrobat/how-to/delete-pages-from-pdf.html#:~:text=Choose%20%E2%80%9CTools%E2%80%9D%20%3E%20%E2%80%9COrganize,or%20pages%20from%20the%20file.}{Adobe Acrobat} (on all OSs), as well as \href{https://superuser.com/questions/517986/is-it-possible-to-delete-some-pages-of-a-pdf-document}{command line tools}.

\end{document}